# Effective Confidence Region Prediction Using Probability Forecasters

David Lindsay [1] and Siân Lindsay [2]

[1] AlgoLabs

david@algolabs.com

AlgoLabs [2]

sian@algolabs.com

**Abstract.** Confidence region prediction is a practically useful extension to the commonly studied pattern recognition problem. Instead of predicting a single label, the constraint is relaxed to allow prediction of a subset of labels given a desired confidence level $1 - \delta$. Ideally, effective region predictions should be (1) well calibrated - predictive regions at confidence level $1 - \delta$ should err with relative frequency at most $\delta$ and (2) be as narrow (or certain) as possible. We present a simple technique to generate confidence region predictions from conditional probability estimates (probability forecasts). We use this 'conversion' technique to generate confidence region predictions from probability forecasts output by standard machine learning algorithms when tested on 15 multi-class datasets. Our results show that approximately 44% of experiments demonstrate well-calibrated confidence region predictions, with the $K$-Nearest Neighbour algorithm tending to perform consistently well across all data. Our results illustrate the practical benefits of effective confidence region prediction with respect to medical diagnostics, where guarantees of capturing the true disease label can be given.

## 1 Introduction

Pattern recognition is a well studied area of machine learning, however the equally useful extension of confidence region prediction [1] has to date received little attention from the machine learning community. Confidence region prediction relaxes the traditional constraint on the learner to predict a single label to allow prediction of a subset of labels given a desired confidence level $1 - \delta$. The traditional approach to this problem was to construct region predictions from so-called $p$-values, often generated from purpose built algorithms such as the Transductive Confidence Machine [1]. Whilst these elegant solutions offer non-asymptotic provenly valid properties, there are significant disadvantages of $p$-values as their interpretation is less direct than that of *probability forecasts* and they are easy to confuse with probabilities. Probability forecasting is an increasingly popular generalisation of the standard pattern recognition problem; rather than attempting to find the "best" label, the aim is to estimate the conditional probability (otherwise known as a probability forecast) of a possible label given an observed object. Meaningful probability forecasts are vital when learning techniques are involved with cost sensitive decision-making (as in medical diagnostics) [2],[3]. In this study we introduce a simple, yet provably valid technique to generate confidence region predictions from probability forecasts, in order to capitalise on the individual benefits of each classification methodology. There are two intuitive criteria for assessing the effectiveness of region predictions, ideally they should be:

1. **Well calibrated** - Predictive regions at confidence level $1-\delta \in [0,1]$ will be wrong (not capture the true label) with relative frequency at most $\delta$.
2. **Certain** - The predictive regions should be as narrow as possible.

The first criterion is the priority: without it, the meaning of predictive regions is lost, and it becomes easy to achieve the best possible performance. We have applied our simple conversion technique to the probability forecasts generated by several commonly used machine learning algorithms on 15 standard multi-class datasets. Our results have demonstrated that 44% of experiments produced well-calibrated region predictions, and found at least one well-calibrated region predictor for 12 of the 15 datasets tested. Our results have also demonstrated our main dual motivation that the learners also provide useful probability forecasts for more direct decision making. We discuss the application of this research with respect to medical diagnostics, as several clinical studies have shown that the use of computer-based diagnostics systems [4], [5] can provide valuable insights into patient problems, especially with helping to identify alternative diagnoses. We argue that with the use of effective probability forecasts and confidence region predictions on multi-class diagnostic problems, the end user can have probabilistic guarantees on narrowing down the true disease. Finally, we identify areas of future research.

## 2 Using Probability Forecasts for Region Prediction

Our notation will extend upon the commonly used supervised learning approach to pattern recognition. Nature outputs information pairs called *examples*. Each example $z_i = (\mathbf{x}_i, y_i) \in \mathbf{X} \times \mathbf{Y} = \mathbf{Z}$ consists of an *object* $\mathbf{x}_i \in \mathbf{X}$ and its *label* $y_i \in \mathbf{Y} = \{1, 2, \ldots, |\mathbf{Y}|\}$. Formally our learner is a function $\Gamma : \mathbf{Z}^n \times \mathbf{X} \to \mathbf{P}(\mathbf{Y})$ of $n$ training examples, and a new test object mapping onto a probability distribution over labels. Let $\hat{P}(y_i = j \mid \mathbf{x}_i)$ represent the estimated conditional probability of the $j$th label matching the true label for the $i$th object tested. We will often consider a sequence of $N$ probability forecasts for the $|\mathbf{Y}|$ possible labels output by a learner $\Gamma$, to represent our predictions for the test set of objects, where the true label is withheld from the learner.

### 2.1 Probability Forecasts Versus $p$-Values

Previous studies [1] have shown that it is possible to create special purpose learning methods to produce valid and asymptotically optimal region predictions via the use of $p$-values. However these $p$-values have several disadvantages when compared to probability forecasts; their interpretation is less direct than that of probability forecasts, the $p$-values do not sum to one (although they often take values between 0 and 1) and they are easy to confuse with probabilities. This misleading nature of $p$-values and other factors have led some authors [6] to object to any use of $p$-values.

In contrast, although studies in machine learning tend to concentrate primarily on 'bare' predictions, probability forecasting has become an increasingly popular doctrine. Making *effective* probability forecasts is a well studied problem in statistics and weather forecasting [7][8]. Dawid (1985) details two simple criteria for describing how effective probability forecasts are:

1. **Reliability** - The probability forecasts "should not lie". When a probability $\hat{p}$ is assigned to an label, there should be roughly $1 - \hat{p}$ relative frequency of the label not occurring.
2. **Resolution** - The probability forecasts should be practically useful and enable the observer to easily rank the labels in order of their likelihood of occurring.

These criteria are analogous to that defined earlier for effective confidence region predictions; indeed we argue that the first criterion of reliability should remain the main focus, as the second resolution criterion naturally is ensured by the classification accuracy of learning algorithms (which is a more common focus of empirical studies). At present the most popular techniques for assessing the quality of probability forecasts are *square loss* (a.k.a. Brier score) and *ROC curves* [9]. Recently we developed the *Empirical Reliability Curve* (ERC) as a visual interpretation of the theoretical definition of reliability [3]. Unlike the square loss and ROC plots, the ERC allows visualisation of over- and under-estimation of probability forecasts.

## 2.2 Converting Probability Forecasts into Region Predictions

The first step in converting the probability forecasts into region predictions is to sort the probability forecasts into increasing order. Let $\hat{y}_i^{(1)}$ and $\hat{y}_i^{(|\mathbf{Y}|)}$ denote the labels corresponding to the smallest and largest probability forecasts respectively

$$\hat{P}(\hat{y}_i^{(1)} \mid \mathbf{x}_i) \leq \hat{P}(\hat{y}_i^{(2)} \mid \mathbf{x}_i) \leq \ldots \leq \hat{P}(\hat{y}_i^{(|\mathbf{Y}|)} \mid \mathbf{x}_i) \ .$$

Using these ordered probability forecasts we can generate the region prediction $\Gamma^{1-\delta}$ at a specific confidence level $1 - \delta \in [0, 1]$ for each test example $\mathbf{x}_i$

$$\Gamma^{1-\delta}(\mathbf{x}_i) = \{\hat{y}_i^{(j)}, \ldots, \hat{y}_i^{(|\mathbf{Y}|)}\} \text{ where } j = \operatorname*{argmax}_k \left\{ \sum_{l=1}^{k-1} \hat{P}(\hat{y}_i^{(l)} \mid \mathbf{x}_i) < \delta \right\} \ . \quad (1)$$

Intuitively we are using the fact that the probability forecasts are estimates of conditional probabilities and we assume that the labels are mutually exclusive. Therefore summing these forecasts becomes a conditional probability of a conjunction of labels, which can be used to choose the labels to include in the confidence region predictions at the desired confidence level.

## 2.3 Assessing the Quality of Region Predictions

Now that we have specified how region predictions are created, how do we assess the quality of them? Taking inspiration from [1], informally we want two criteria to be satisfied: (1) that the region predictions are well-calibrated, i.e. the fraction of errors at confidence level $1 - \delta$ should be at most $\delta$, and (2) that the regions should be as narrow as possible. More formally let $\mathrm{err}_i(\Gamma^{1-\delta}(\mathbf{x}_i))$ be the error incurred by the true label $y_i$ not being in the region prediction of object $\mathbf{x}_i$ at confidence level $1-\delta$

$$\mathrm{err}_i(\Gamma^{1-\delta}(\mathbf{x}_i)) = \begin{cases} 1 \text{ if } y_i \notin \Gamma^{1-\delta}(\mathbf{x}_i) \\ 0 \text{ otherwise} \end{cases} \quad . \tag{2}$$

To assess the second criteria of uncertainty we simply measure the fraction of total labels that are predicted at the desired confidence level $1-\delta$

$$\mathrm{unc}_i(\Gamma^{1-\delta}(\mathbf{x}_i)) = \frac{|\Gamma^{1-\delta}(\mathbf{x}_i)|}{|\mathbf{Y}|} \quad . \tag{3}$$

These terms can then be averaged over all the $N$ test examples

$$\mathrm{Err}_N^{1-\delta} = \frac{1}{N}\sum_{i=1}^{N}\mathrm{err}_i(\Gamma^{1-\delta}(\mathbf{x}_i)) \, , \, \mathrm{Unc}_N^{1-\delta} = \frac{1}{N}\sum_{i=1}^{N}\mathrm{unc}_i(\Gamma^{1-\delta}(\mathbf{x}_i)) \quad . \tag{4}$$

It is possible to prove that this technique of converting probability forecasts into region predictions (given earlier in Equation 1) is well-calibrated for a learner that outputs Bayes optimal forecasts, as shown below.

**Theorem 1.** *Let $\boldsymbol{P}$ be a probability distribution on $\mathbf{Z} = \mathbf{X} \times \mathbf{Y}$. If a learners forecasts are Bayes optimal i.e. $\hat{P}(y_i = j \mid \mathbf{x}_i) = \boldsymbol{P}(y_i = j \mid \mathbf{x}_i), \forall j \in \mathbf{Y}, 1 \leq i \leq N$, then*

$$\mathbf{P}^N\left(\mathrm{Err}_N^{1-\delta} \geq \delta + \varepsilon\right) \leq e^{-2\varepsilon^2 N} \quad .$$

This means if the probability forecasts are Bayes optimal, then even for small finite sequences the confidence region predictor will be well-calibrated with high probability. This proof follows directly from the Hoeffding Azuma inequality [10], and thus justifies our method for generating confidence region predictions.

## 2.4    An Illustrative Example: Diagnosis of Abdominal Pain

In this example we illustrate our belief that both probability forecasts *and* region predictions are practically useful. Table 1 shows probability forecasts for all possible disease labels made by Naive Bayes and DW13-NN learners when tested on the Abdominal Pain dataset from Edinburgh Hospital, UK [5]. For comparison, the predictions made by both learners are given for the same patient examples in the Abdominal Pain dataset. At a glance it is obvious that the predicted probabilities output by the Naive Bayes learner are far more extreme (i.e. very close to 0 or 1) than those output by the DW13-NN learner. Example 1653 (Table 1 rows 1 and 4) shows a patient object which is predicted correctly by the Naive Bayes learner ($\hat{p}$=0.99) and less emphatically by the DW13-NN learner ($\hat{p}$=0.85).

  Example 2490 demonstrates the problem of over- and under- estimation by the Naive Bayes learner where a patient is incorrectly diagnosed with Intestinal obstruction (overestimation), yet the true diagnosis of Dyspepsia is ranked 6th with a very low predicted probability of $\hat{p} = \frac{22}{1000}$ (underestimation). In contrast the DW13-NN learner makes more *reliable* probability forecasts; for example 2490, the true class is correctly predicted albeit with lower predicted probability. Example 5381 demonstrates a situation where both learners encounter an error in their predictions. The Naive Bayes learner gives misleading predicted probabilities of $\hat{p} = 0.969$ for the incorrect diagnosis of Appendicitis, and a mere $\hat{p} = 0.03$ for the true class label of Non-specific abdominal pain. In contrast, even though the DW13-NN learner incorrectly predicts Appendicitis, it is with far less certainty $\hat{p} = 0.53$ and if the user were to look at all probability forecasts it would be clear that the true class label should not be ignored with a predicted probability $\hat{p} = 0.425$.

**Table 1.** This table shows sample predictions output by the Naive Bayes and DW13-NN learners tested on the Abdominal Pain dataset. The 9 possible class labels for the data (left to right in the table) are: Appendicitis, Diverticulitis, Perforated peptic ulcer, Non-specific abdominal pain, Cholisistitis, Intestinal obstruction, Pancreatitis, Renal colic and Dyspepsia. The associated probability of the predicted label is underlined and emboldened, whereas the actual class label is marked with the ✠ symbol

| Example # | Probability forecast for each class label | | | | | | | | |
|---|---|---|---|---|---|---|---|---|---|
| | Appx. | Div. | Perf. | Non-spec. | Cholis. | Intest. | Pancr. | Renal | Dyspep. |
| **DW13-NN** | | | | | | | | | |
| 1653 | 0.0 | 0.0 | 0.0 | 0.03 | **0.85**✠ | 0.0 | 0.01 | 0.0 | 0.11 |
| 2490 | 0.0 | 0.0 | 0.22 | 0.0 | 0.0 | 0.25 | 0.04 | 0.09 | **0.4**✠ |
| 5831 | **0.53** | 0.0 | 0.0 | 0.425✠ | 0.001 | 0.005 | 0.0 | 0.0 | 0.039 |
| **Naive Bayes** | | | | | | | | | |
| 1653 | 3.08e-9 | 4.5e-6 | 3.27e-6 | 4.37e-5 | **0.99**✠ | 4.2e-3 | 3.38e-3 | 4.1e-10 | 1.33e-4 |
| 2490 | 9.36e-5 | 0.01 | 0.17 | 2.26e-5 | 0.16 | **0.46** | 0.2 | 2.17e-7 | 2.2e-4✠ |
| 5831 | **0.969** | 2.88e-4 | 1.7e-13 | 0.03✠ | 1.33e-9 | 2.2e-4 | 4.0e-11 | 6.3e-10 | 7.6e-9 |

The results below show region predictions output at 95% and 99% confidence levels by the Naive Bayes and DW13-NN learners for the same Abdominal Pain examples tested in Table 1 using the simple conversion technique (cf. 1). As before, the true label is marked with the ✠ symbol if it is contained in the region prediction.

**Conversion of DW13-NN forecasts into region predictions:**

| Example # | Region at 95% Confidence | Region at 99% Confidence |
|---|---|---|
| 1653 | $\{$ Cholis ✠, Dyspep $\}$ | $\{$ Cholis ✠, Dyspep, Non-spec, Pancr $\}$ |
| 2490 | $\{$ Dyspep ✠, Intest, Perf, Renal $\}$ | $\{$ Dyspep ✠, Intest, Perf, Renal, Pancr $\}$ |
| 5831 | $\{$ Appx, Non-spec ✠ $\}$ | $\{$ Appx, Non-spec ✠, Dyspep $\}$ |

**Conversion of Naive Bayes forecasts into region predictions:**

| Example # | Region at 95% Confidence | Region at 99% Confidence |
|---|---|---|
| 1653 | $\{$ Cholis ✠ $\}$ | $\{$ Cholis ✠ $\}$ |
| 2490 | $\{$ Intest, Pancr, Perf, Cholis $\}$ | $\{$ Intest, Pancr, Perf, Cholis, Div $\}$ |
| 5831 | $\{$ Appx $\}$ | $\{$ Appx, Non-spec ✠ $\}$ |

The DW13-NN learner's region predictions are wider, with a total of 20 labels included in all region predictions as opposed to 14 for the Naive Bayes learner. However, despite these narrow predictions, the Naive Bayes's region prediction fails to capture the true label for example 2490, and only captures the true label for example 5831 at a 99% confidence level. In contrast, the DW13-NN's region predictions successfully capture each true label even at a 95% confidence level. This highlights what we believe to be the practical benefit of generating well-calibrated region predictions - the ability to capture or dismiss labels with the guarantee of being correct is highly desirable. In our example, narrowing down the possible diagnosis of abdominal pain could save money and patient anxiety in terms of unnecessary tests to clarify the condition of the patient. It is important to stress that when working with probabilistic classification, classification accuracy should not be the only mechanism of assessment. Table 2 shows that the Naive Bayes learner actually has a slightly lower classification error rate of 29.31% compared with 30.86% of the DW13-NN learner. If other methods of learner assessment had been ignored, the Naive Bayes learner may have been used for prediction by the physician even though Naive Bayes has less reliable probability forecasts and non-calibrated region predictions.

## 2.5    The Confidence Region Calibration (CRC) Plot Visualisation

The Confidence Region Calibration (CRC) plot is a simple visualisation of the performance criteria detailed earlier (cf. 4). Continuing with our example, figure 1 shows CRC plots for the Naive Bayes and DW13-NN's probability forecasts on the Abdominal pain dataset. The CRC plot displays all possible confidence levels on the horizontal axis, versus the total fraction of objects or class labels on the vertical axis. The dashed line represents the number of errors $\text{Err}_N^{1-\delta}$, and the solid line represents the average region width $\text{Unc}_N^{1-\delta}$ at each confidence level $1 - \delta$ for all $N$ test examples. Informally, as we increase the confidence level, the errors should decrease and the width of region predictions increase. The lines drawn on the CRC plot give the reader at a glance useful information about the learner's region predictions. If the error line never deviates above the upper left-to-bottom right diagonal (i.e. $\text{Err}_N^{1-\delta} \leq \delta$ for all $\delta \in [0, 1]$), then the learner's region predictions are *well-calibrated*. The CRC plot also provides a useful measure of the quality of the probability forecasts, we can check if regions are well-calibrated by computing the area between the main diagonal $1 - \delta$ and the error line $\text{Err}_n^{1-\delta}$, and the area under the average region width line $\text{Unc}_n^{1-\delta}$ using simple trapezium and triangle rule estimation (see Table 2). These deviation areas (also seen beneath Figure 1) can enable the user to check the performance of region predictions without any need to check the corresponding CRC plot.

Figure 1 shows that in contrast to the DW13-NN learner, the Naive Bayes learner is not well-calibrated as demonstrated by its large deviation above the left-to-right diagonal. At a 95% confidence level, the Naive Bayes learner makes 18% errors on its confidence region predictions and not $\leq 5\%$ as is usually required. The second line corresponding to the average region width $\text{Unc}_n^{1-\delta}$ shows how useful the learner's region predictions are. The Naive Bayes CRC plot indicates that at 95% confidence, roughly 15% of the 9 possible disease labels ($\approx 1.4$) are predicted. In contrast, the DW13-NN learner is less certain at this confidence level, predicting 23% of labels ($\approx 2.1$). However, although the DW13-NN learner's region predictions are slightly wider than those of Naive Bayes, the DW13-NN learner is far closer to being calibrated, which is our first priority. Indeed narrowing down the true disease of a patient from 9 to 2 possible disease class labels, whilst guaranteeing less that 5% chance of errors in that region prediction, is a desirable goal. In addition, further information could be provided by referring back to the original probability forecasts to achieve a more direct analysis of the likely disease.

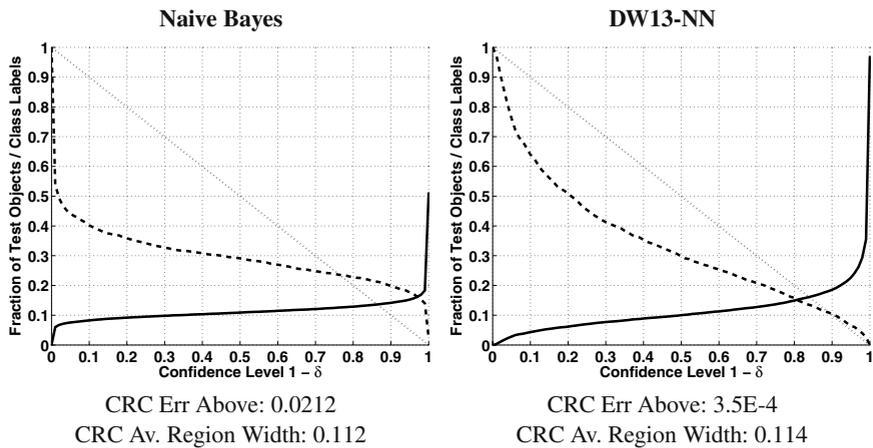

**Fig. 1.** Confidence Region Calibration (CRC) plots of probability forecasts output by Naive Bayes and DW 13-NN learners on the Abdominal Pain dataset. The dashed line represents the fraction of errors made on the test sets region predictions at each confidence level. The solid line represents the average region width for each confidence level $1 - \delta$. If the dashed error line never deviates above the main left-right diagonal then the region predictions are well-calibrated. The deviation areas of both error and region width lines are given beneath each plot

## 3   Experimental Results

We tested the following algorithms as implemented by the WEKA Data Mining System [9]: Bayes Net, Neural Network, Distance Weighted $K$-Nearest Neighbours (DW$K$-NN), C4.5 Decision Tree (with Laplace smoothing), Naive Bayes and Pairwise Coupled + Logistic Regression Sequential Minimisation Optimisation (PC+LR SMO) a computationally efficient version of Support Vector Machines. We tested the following 15 multi-class datasets (number of examples, number of attributes, number of classes): Abdominal Pain (6387,136,9) from [5], and the other datasets from the UCI data repository [11]: Anneal (898,39,5), Glass (214,10,6), Hypothyroid (3772,30,4), Iris (150,5,3), Lymphography (148,19,4), Primary Tumour (339,18,21), Satellite Image (6435,37,6), Segment (2310,20,7), Soy Bean (683,36,19), Splice (3190,62,3), USPS Handwritten Digits (9792,256,10), Vehicle (846,19,4), Vowel (990,14,11) and Waveform ( 0,41,3). Each experiment was carried out on 5 randomisations of a 66% Training / 34% Test split of the data. Data was normalised, and missing attributes replaced with mean and mode values for numeric and nominal attributes respectively. Deviation areas for the CRC plots were estimated using 100 equally divided intervals. The stan-

**Table 2.** Results of CRC Error Above / CRC Av. Region Width (assessing confidence region predictions), Classification Error Rate / Square Loss (assessing probability forecasts), computed from the probability forecasts of several learners on 15 multi-class datasets. Experiments that gave well-calibrated confidence region predictions (where CRC Error Above = 0) are highlighted in grey. Best assessment score of the well-calibrated learners (where available) for each dataset are emboldened

**CRC Plot Error Above / CRC Plot Av. Region Width**

| Dataset / Learner | Naive Bayes | Bayes Net | Neural Net | C4.5 | PC+LR SMO | DW $K$-NN |
|---|---|---|---|---|---|---|
| Abdominal Pain | 0.021 / 0.112 | - / - | 0.027 / **0.112** | 0.024 / 0.155 | 0.002 / 0.113 | **3.5E-4** / 0.114 |
| Anneal | 0.004 / 0.167 | 0.0 / **0.167** | 0.0 / 0.169 | 0.0 / 0.176 | 0.0 / 0.168 | 0.0 / 0.169 |
| Glass | 0.098 / **0.144** | 0.005 / 0.145 | 0.025 / 0.145 | 0.017 / 0.202 | 0.02 / 0.144 | **0.004** / 0.146 |
| Hypothyroid | 0.0 / **0.250** | 0.0 / 0.251 | 1.6E-4 / 0.252 | 0.0 / 0.253 | 0.0 / 0.251 | 0.0 / 0.252 |
| Iris | 0.0 / **0.332** | 1.2E-4 / 0.335 | 1.6E-4 / 0.335 | 7.5E-4 / 0.339 | 0.0 / 0.334 | 0.0 / 0.335 |
| Lymphography | 0.002 / 0.251 | 2.8E-4 / 0.252 | 0.006 / 0.252 | 0.004 / 0.284 | 0.002 / 0.252 | 0.0 / **0.251** |
| Primary Tumour | 0.146 / **0.048** | **0.133** / 0.048 | 0.156 / 0.048 | 0.219 / 0.098 | 0.153 / 0.047 | 0.24 / 0.048 |
| Satellite Image | 0.018 / 0.166 | - / - | 0.002 / 0.168 | 0.0 / 0.189 | 0.0 / **0.169** | 0.0 / 0.169 |
| Segment | 0.012 / 0.143 | 0.0 / **0.143** | 0.0 / 0.145 | 0.0 / 0.154 | 0.0 / 0.145 | 0.0 / 0.146 |
| Soy Bean | 0.001 / 0.053 | 0.0 / **0.053** | 5.5E-4 / 0.056 | 1.7E-4 / 0.1055 | 2.9E-4 / 0.055 | 0.0 / 0.057 |
| Splice | 0.0 / **0.334** | - / - | 1.2E-4 / 0.333 | 0.0 / 0.344 | 0.001 / 0.334 | 0.0 / 0.334 |
| USPS | 0.018 / 0.099 | - / - | 0.0 / **0.102** | 0.0 / 0.121 | 0.0 / 0.102 | 0.0 / 0.104 |
| Vehicle | 0.083 / 0.251 | 0.0 / 0.251 | 0.004 / 0.25142 | 0.002 / 0.274 | 0.0 / **0.251** | 0.0 / 0.252 |
| Vowel | 0.006 / 0.093 | 2.7E-4 / 0.094 | 0.001 / 0.093 | 0.002 / 0.153 | 0.002 / 0.093 | 0.0 / **0.094** |
| Waveform    0 | 0.006 / 0.334 | 0.0 / **0.334** | 0.006 / 0.332 | 0.004 / 0.348 | 0.0 / 0.335 | 0.0 / 0.335 |

**Error Rate / Square Loss**

| Dataset / Learner | Naive Bayes | Bayes Net | Neural Net | C4.5 | PC+LR SMO | DW $K$-NN |
|---|---|---|---|---|---|---|
| Abdominal Pain | 29.309 / 0.494 | - / - | 30.484 / 0.528 | 38.779 / 0.597 | **26.886 / 0.388** | 30.859 / 0.438 |
| Anneal | 11.929 / 0.219 | 4.794 / 0.076 | **1.672 / 0.026** | 1.895 / 0.051 | 2.676 / 0.044 | 2.899 / 0.06 |
| Glass | 56.056 / 0.869 | **33.521 / 0.47** | 37.747 / 0.58 | 35.775 / 0.579 | 40.282 / 0.537 | 37.465 / 0.509 |
| Hypothyroid | 4.423 / 0.075 | 1.002 / 0.016 | 5.521 / 0.095 | **0.461 / 0.009** | 2.975 / 0.046 | 6.858 / 0.118 |
| Iris | 4.8 / 0.089 | 6.8 / 0.1 | 5.6 / 0.089 | 7.2 / 0.136 | 4.8 / 0.068 | **3.2 / 0.057** |
| Lymphography | 16.735 / 0.279 | 15.51 / 0.248 | 20.0 / 0.346 | 22.857 / 0.368 | 22.041 / 0.317 | **20.408 / 0.29** |
| Primary Tumour | 53.982 / 0.692 | 54.867 / **0.695** | 59.646 / 0.905 | 57.699 / 0.855 | 58.23 / 0.772 | **53.628** / 0.721 |
| Satellite Image | 20.793 / 0.4 | - / - | 10.839 / 0.19 | 14.965 / 0.249 | 14.382 / 0.194 | **9.604 / 0.138** |
| Segment | 19.403 / 0.357 | 4.364 / 0.071 | 3.688 / **0.061** | 4.156 / 0.078 | 7.273 / 0.107 | **3.61** / 0.066 |
| Soy Bean | 9.604 / 0.163 | **6.432 / 0.091** | 7.489 / 0.125 | 7.753 / 0.263 | 8.37 / 0.134 | 9.779 / 0.142 |
| Splice | **4.365 / 0.068** | - / - | 4.774 / 0.079 | 6.529 / 0.116 | 7.267 / 0.133 | 16.764 / 0.269 |
| USPS | 19.619 / 0.388 | - / - | **3.055 / 0.054** | 11.    / 0.201 | 4.679 / 0.077 | 3.162 / 0.055 |
| Vehicle | 53.192 / 0.848 | 29.787 / 0.389 | 20.213 / 0.319 | 27.801 / 0.403 | **26.312 / 0.363** | 29.078 / 0.388 |
| Vowel | 40.606 / 0.538 | 21.455 / 0.308 | 9.152 / 0.155 | 24.303 / 0.492 | 34.909 / 0.452 | **1.697 / 0.034** |
| Waveform    0 | 20.12 / 0.342 | 18.619 / 0.268 | 16.627 / 0.297 | 24.658 / 0.407 | **13.974 / 0.199** | 19.328 / 0.266 |

dard inner product was used in all SMO experiments. All experiments were carried out on a Pentium 4 1.7Ghz PC with 512Mb RAM.

The values given in Table 2 are the average values computed from each of the 5 repeats of differing randomisations of the data. The standard deviation of these values was negligible of magnitude between E-8 and E-12. It is important to note that all results are given with the default parameter settings of each algorithm as specified by WEKA - there was no need for an exhaustive search of parameters to achieve well-calibrated region predictions. Results with the DW$K$-NN learners are given for the best value of $K$ for each dataset, which are $K$ = 13, 2, 9, 11, 5, 6, 12, 7, 2, 3, 6, 4, 9, 1, 10 from top to bottom as detailed in Table 2. These choices of $K$ perhaps reflect some underlying noise level or complexity in the data, so a higher value of $K$ corresponds to a noisier, more complex dataset. Results with Bayesian Belief Networks were unavailable due to mem-ory constraints with the following datasets: Abdominal Pain, Satellite Image, Splice and USPS. Results which produced well-calibrated region predictions are highlighted in grey, where 38 out of 86 experiments ($\approx$44%) were calibrated. Figure 1 illustrates that despite DW13-NN not being strictly well-calibrated on the Abdominal Pain dataset with a CRC error above calibration $\approx$3.5E-4, in practice this deviation appears negligible. Indeed, if we loosened our definition of well-calibratedness to small deviations of order $\approx$E-4 then the fraction of well-calibrated experiments would increase to 49 out of 86 ($\approx$57%). All datasets, except Abdominal Pain, Glass and Primary Tumor were found to have at least one calibrated learner, and all learners were found to be well-calibrated for at least one dataset. Some datasets, such as Anneal and Segment, were found to have a choice of 5 calibrated learners. In this instance it is important to consult other performance criteria (i.e. region width or classification error rate) to assess the quality of the probability forecasts and decide which learner's predictions to use. Table 2 shows that the Naive Bayes and Bayes Net confidence region predictions are far narrower ($\approx -4$%) than the other learners, however this may explain the Naive Bayes learner's poor calibration performance across data. In contrast the C4.5 Decision Tree learner has the widest of all region predictions ($\approx +13$%). The Bayes Net performs best in terms of square loss, whilst the Naive Bayes performs the worst. This reiterates conclusions found in other studies [3], in addition to results in Section 2.4 where the Naive Bayes learner's probability forecasts were found to be unreliable. Interestingly, the $K$-NN learners give consistently well-calibrated region predictions across datasets - perhaps this behaviour can be attributed to the $K$-NN learners' guarantee to converge to the Bayes optimal probability distribution [10].

## 4    Discussion

We believe that using both probability forecasts and region predictions can give the end-users of such learning systems a wealth of information to make effective decisions in cost-sensitive problem domains, such as in medical diagnostics. Medical data often contains a lot of noise and so high classification accuracy is not always possible (as seen here with Abdominal Pain data, see Table 2). In this instance the use of probability forecasts and region predictions can enable the physician to filter out improbable diagnoses, and so choose the best course of action for each patient with safe knowledge of their likelihood of error. In this study we have presented a simple methodology to convert probability forecasts into confidence region predictions. We were surprised at how successful standard learning techniques were at solving this task. It is interesting to consider why it was not possible to obtain exactly calibrated results with the Abdominal Pain, Glass and Primary Tumor datasets. Further research is being conducted to see if meta-learners such as Bagging [12] and Boosting [13] could be used to improve on these initial results and obtain better calibration for these problem datasets.

As previously stated, the Transductive Confidence Machine (TCM) framework has much empirical and theoretical evidence for strong region calibration properties, however these properties are derived from the $p$-values which can be misleading if read on their own, as they do not have any direct probabilistic interpretation. Indeed, our results have shown that unreliable probability forecasts can be misleading enough, but with $p$-values this confusion is compounded. Our current research is to see if the exposition of this study could be done in reverse; namely a method could be developed to convert $p$-values into effective probability forecasts.

Another interesting extension to this research would be to investigate the problem of multi-label prediction, where class labels are not mutually exclusive [14]. Possibly an adaptation of the confidence region prediction method could be applied for this problem domain. Much of the current research in multi-label prediction is concentrated in document and image classification, however we believe that multi-label prediction in medical diagnostics is an equally important avenue of research, and particularly relevant in instances where a patient may suffer more than one illness at once.

## Acknowledgements

We would like to acknowledge Alex Gammerman and Volodya Vovk for their helpful suggestions, and in reviewing our work. This work was supported by EPSRC (grants GR46670, GR/P01311/01), BBSRC (grant B111/B1014428), and MRC (grant S505/65).